\documentclass[sigconf]{acmart}

\AtBeginDocument{%
  \providecommand\BibTeX{{%
    \normalfont B\kern-0.5em{\scshape i\kern-0.25em b}\kern-0.8em\TeX}}}

\usepackage{algorithm}
\usepackage{algorithmic}
\begin{document}
\copyrightyear{2019}
\acmYear{2019}
\setcopyright{acmcopyright}
\acmConference[DRL4KDD '19]{The 1st Workshop on Deep Reinforcement Learning for Knowledge Discovery}{August 5, 2019}{Anchorage, AK, USA}

\acmBooktitle{The 1st Workshop on Deep Reinforcement Learning for Knowledge Discovery (DRL4KDD '19), August 5, 2019, Anchorage, AK, USA}

\title{Reinforcement Learning Driven Heuristic Optimization}

\author{Qingpeng Cai}
\email{cqp14@mails.tsinghua.edu.cn}
\affiliation{%
  \institution{Tsinghua University}
  \city{Beijing}
  \country{China}}

\author{Will Hang}
\email{willhang@stanford.edu}
\affiliation{%
  \institution{Stanford University}
  \city{Stanford}
  \state{California}}

\author{Azalia Mirhoseini}
\email{azalia@google.com}
\affiliation{%
  \institution{Google Brain}
  \city{Mountain View}
  \state{California}}

\author{George Tucker}
\email{gjt@google.com}
\affiliation{%
  \institution{Google Brain}
  \city{Mountain View}
  \state{California}}
  
\author{Jingtao Wang}
\email{jingtaow@google.com}
\affiliation{%
  \institution{Google AI}
  \city{Beijing}
  \country{China}}

\author{Wei Wei}
\email{wewei@google.com}
\affiliation{%
  \institution{Google AI}
  \city{Sunnyvale}
  \state{California}}

\renewcommand{\shortauthors}{Q. Cai et al.}

\begin{abstract}
Heuristic algorithms such as simulated annealing, Concorde, and METIS are effective and widely used approaches to find solutions to combinatorial optimization problems. However, they are limited by the high sample complexity required to reach a reasonable solution from a cold-start. In this paper, we introduce a novel framework to generate better initial solutions for heuristic algorithms using reinforcement learning (RL), named RLHO. We augment the ability of heuristic algorithms to greedily improve upon an existing initial solution generated by RL, and demonstrate novel results where RL is able to leverage the performance of heuristics as a learning signal to generate better initialization. 

We apply this framework to Proximal Policy Optimization (PPO) and Simulated Annealing (SA). We conduct a series of experiments on the well-known NP-complete bin packing problem, and show that the RLHO method outperforms our baselines. We show that on the bin packing problem, RL can learn to help heuristics perform even better, allowing us to combine the best parts of both approaches.
\end{abstract}

\ccsdesc[500]{Computing methodologies~Sequential decision making}
\ccsdesc[300]{Computing methodologies~Planning under uncertainty}
\ccsdesc[300]{Computing methodologies~Discrete space search}

\keywords{Deep reinforcement learning, heuristic algorithms, combinatorial optimization}

\maketitle

\section{Introduction}

Combinatorial optimization \cite{wolsey2014integer} aims to find the optimal solution with the minimum cost from a finite set of candidates to discrete problems such as the bin packing problem, the traveling salesman problem, or integer programming. Combinatorial optimization has seen broad applicability in fields ranging from telecommunications network design, to task scheduling, to transportation systems planning. As many of these combinatorial optimization problems are NP-complete, optimal solutions cannot be tractably found \cite{ausiello2012complexity}. 

Heuristic algorithms such as simulated annealing (SA) \cite{rutenbar1989simulated,aarts1988simulated,van1987simulated} are designed to search for the optimal solution by randomly perturbing candidate solutions and accepting those that satisfy some greedy criterion such as Metropolis-Hastings. Heuristics are widely used in combinatorial optimization problems such as Concorde for the traveling salesman problem, or METIS for graph partitioning \cite{concorde, METIS}. Some heuristic algorithms like SA are theoretically guaranteed to find the optimal solution to a problem given a low enough temperature and enough perturbations \cite{Ingber:1993:SAP:2262405.2262615}.


However, the framework for heuristic algorithms begins the solution search from a randomly initialized candidate solution. For example, in the bin packing problem, the initial solution fed into SA would be a random assignment of objects to bins, which would then be repeatedly perturbed until convergence. Starting hill climbing from a cold start is time-consuming and limits the applicability of heuristic algorithms on practical problems.

Reinforcement learning (RL) has been proposed as a technique to yield efficient solutions to combinatorial optimization problems by first learning a policy, and then using it to generate a solution to the problem. RL has seen interesting applications in real world combinatorial optimization problems \cite{DBLP:journals/corr/ZophL16, MirhoseiniPLSLZ17}. However, RL lacks the theoretical guarantees of algorithms like SA, which use a hill-climbing approach and are less susceptible to problems like policy collapse. By setting the greedy criterion to only accept better solutions, SA can achieve monotonically better performance, whereas RL cannot.

Thus, it is best to generate an initial solution using RL and continuously improve this solution using heuristic algorithms like SA. Furthermore, it is advantageous for RL to learn how to provide an optimal initialization to SA to maximize the performance of both techniques in tandem. 

\begin{figure}[]
    \centering
            \includegraphics[scale=0.36]{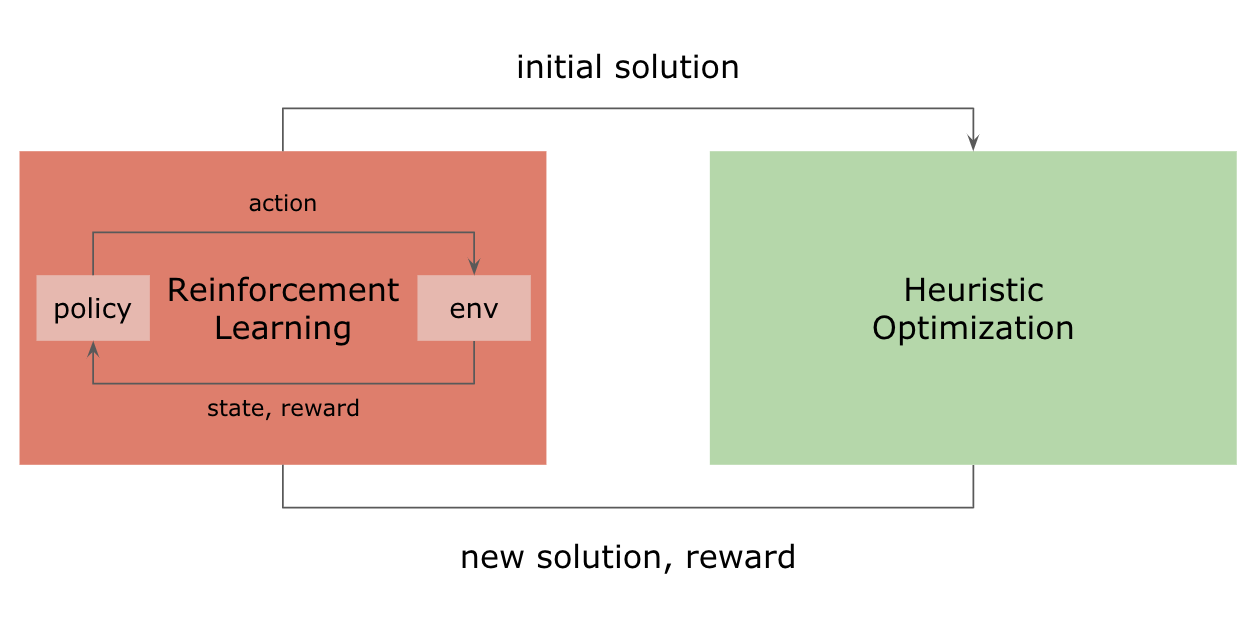}
    \caption{The RLHO framework.}
    \label{fig: framework}
\end{figure}

In this paper, we address these two points by introducing the Reinforcement Learning Driven Heuristic Optimization Framework (RLHO), shown in Figure \ref{fig: framework}. There are two components in this framework: the RL agent and the heuristic optimizer (HO). The RL agent generates solutions that act as initialization for HO, and HO searches for better solutions starting from the solution generated by RL.  After HO finishes executing (upon convergence or after a set number of search steps), it returns the found solution and the reward to the RL agent. Our learning process is an alternating loop of (1) generating initial solutions with RL and then (2) searching for better solutions with HO. To the RL agent, HO is part of the environment.

We apply RLHO to the bin packing problem where the RL agent is modeled using Proximal Policy Optimization (PPO) \cite{schulman2017proximal} and HO is simulated annealing (SA). We demonstrate that not only does combining PPO and SA yield superior performance to PPO alone, but also that PPO is actually able to learn to generate better initialization for SA. By observing the end performance of SA on a problem, PPO can generate inputs to SA that improve the performance of SA itself.

In summary, our contributions in this paper are as follows:

\begin{itemize}
\item We demonstrate a novel approach to combinatorial optimization where reinforcement learning and heuristic algorithms are combined to yield superior results to reinforcement learning alone on a combinatorial optimization problem.

\item We demonstrate that we can train reinforcement learning to enable heuristic algorithms to achieve superior performance than when they are decoupled on a combinatorial optimization problem.

\end{itemize}

\subsection{Related Work}
Reinforcement learning and evolutionary algorithms achieve competitive performance on MuJoCo tasks and Atari games \cite{salimans2017evolution}. The idea of applying evolutionary algorithms to reinforcement learning \cite{moriarty1999evolutionary} has been widely studied.  \cite{khadka2018evolution} proposes a framework to apply evolutionary strategies to selectively mutate a population of reinforcement learning policies. \cite{maheswaranathan2018guided, pourchot2018cem} use a gradient method to enhance evolution.

Our work is different from the above as we apply deep reinforcement learning to generate better initializations for heuristic algorithms. The heuristic part in the RLHO framework only changes the solution, rather than the parameters of the policy. To our knowledge, our work is the first that does this.

\section{Combining PPO and SA}

\subsection{Preliminary Discussion}

What is the best way to combine an RL agent and a heuristic algorithm? A first approach is to allow an RL agent to generate an initial solution to a combinatorial optimization problem, then execute a heuristic algorithm to refine this initial solution until convergence, and then train the RL policy with the rewards obtained from the performance of the heuristic algorithm. This would delineate one episode. However, on large problems, heuristics take a long time to converge. Thus, in our approach, we allow the heuristic algorithm to run for a only limited number of steps in one episode.

We now introduce the RLHO algorithm.

\subsection{The RLHO Algorithm}

Our approach is a two-stage process as detailed in Algorithm \ref{alg:RLHO}: at the start of each episode, first run RL for $x$ steps to generate an initial solution $s_x$. Then, run pure HO for $y$ steps starting from $s_x$. Finally we update RL with the cost of the final solution. We repeat this process with a fresh start every time.

Our action space is designed as perturbing the currently available solution. In our bin packing problem discussed in more detail in Section \ref{experimental}, the agent is first presented with a randomly initialized assignment $s_0$ of items to bins. The environment around the bin packing problem will then present the agent with an item $i$. The agent then needs to decide which other item $j$ to swap locations with item $i$ based on the current state.

For the design of the reward function, we define the intermediate reward as the difference between the cost of the previous solution and the cost of the current solution, as the goal is to minimize cost. 

When the agent's action space consists of perturbations, the MDP for the combinatorial optimization problem results in an infinite horizon. We are not privileged with $V(s_{term}) = 0$ that would normally denote the terminal state of the MDP. The agent is free to continue perturbing the state forever, and thus, $V(s_{term})$ is undefined. However, our agents are trained with a finite number of steps $x$, so $V(s_x)$ would normally need to be estimated with a baseline such as a value function. The value function is a poor estimator because it does not accurately estimate the additional expected performance of the agent in the limit of time, because we simply don't possess such data.

To address this, a novelty in our approach is to obtain a better estimate for $V(s_{x})$ using the performance of HO. The additional optimization provided by HO gives us an additional training signal to RL as to how RL actions contribute to the future return provided by HO. Therefore, RL can be trained by two signals in RLHO: (1) the intermediate reward at each RL step, and (2) the discounted future reward provided by HO conditioned on the initialization provided by RL. This approach provides RL with a training signal to generate better initialization for HO.

\begin{equation}\label{value_function}
\begin{split}
V(s_t) &= \sum_{k=t}^{x-1}{\gamma}^{k}r_{k} +  \sum_{k=x}^{\infty}{\gamma}^{k}r_{k}\\
& = \sum_{k=t}^{x-1}{\gamma}^{k}r_{k} + V(s_{x}).
\end{split}
\end{equation}

As shown in Equation \eqref{value_function}, we can replace the infinite horizon term with a stationary, tractable value $V(s_{x})$. We obtain $V(s_{x})$ by running pure HO for y steps starting from $s_x$, and then taking the difference between the cost of $s_x$ and the cost of the final solution $s_{x+y}$ as an estimate for the value of $V(s_{x})$.

\begin{algorithm}[H]
 
  \caption{The RLHO algorithm}
  \begin{algorithmic}
   \STATE Initialize the replay buffer $\mathcal{B}$ and the solution randomly\\
   \STATE Initialize the number of RL steps $x$ and the number of SA steps $y$ in one episode\\
  \FOR{iteration $=1,2, ...$} 
    \STATE Rollout using RL policy for $x$ steps and store the transitions in $\mathcal{B}$, obtaining initial solution $s_x$ from RL
	\STATE Run HO on $s_x$ for $y$ steps to obtain $s_{x + y}$
	\STATE Get the new reward $r_n$ as the difference of costs of $s_{x}$ and $s_{x+y}$\\
    \STATE Train RL using $V(s_x)$\\
    \STATE Reset the solution and hyperparameters of HO
    \ENDFOR
\end{algorithmic}
 \label{alg:RLHO}
\end{algorithm}

\begin{algorithm}[H]
  \caption{Simulated Annealing}
  \begin{algorithmic}
  \STATE Initialize the temperature $T=t_m$, the maximal number of steps of SA in one path, $y$ \\
    \STATE $q=0$, $a=-ln(\frac{t_{m}}{t_{0}})$
    \STATE Obtain the PPO solution $s_x$
    \FOR{$t = 1,2,...,y$}
    \STATE Perturb the current solution $s_{x + t}$ randomly, get $s'_{x + t}$\\
    \IF {$cost(s'_{x + t}) > cost(s_{x + t})$} 
    \STATE Reject $s'_{x + t}$ with probability \\   $p = 1-{e}^{-(c(s')-c(s))/T}$\\
    \ELSE
    \STATE $s_{x + t + 1} = s'_{x + t}$
    \ENDIF
    \STATE $T= t_m {e}^{aq/y}$
    \STATE $q=q+1$
	\ENDFOR
  \end{algorithmic}
 \label{alg:SA}
\end{algorithm}

\section{Performance Evaluation} \label{experimental}

We validate our methods on the bin packing problem. In this section we first introduce the bin packing problem, and then discuss the performance gain obtained when combining the RL part (PPO) and the heuristic optimizer (SA) in our RLHO framework. The details of SA are shown in Algorithm \ref{alg:SA}.

\subsection{The Bin Packing Problem}

Bin packing is a classical combinatorial optimization problem where the objective is to use the minimum number of bins to pack items of different sizes, with the constraint that the sum of sizes of items in one bin is bounded by the size of the bin. Let $n$ denote the number of bins and the number of items, and $v$ denote the vector representing the of sizes of all items. Let $x_{ij}$ be the 0/1 matrix that represents one assignment of items to bins (a packing), i.e., $x_{ij}=1$ means the item $j$ is put in the bin $i$. Given a packing $x$, let $c(x)$ denote the cost, the number of bins used in this solution, i.e., $c(x)=\sum_{i=1}^{n}\sum_{j=1}^{n} x_{ij}$.

\subsection{Learning to Generate Better Initializations}

We evaluate the ability of RLHO to generate better initializations for heuristic algorithms. In this set of experiments, during training, we allow RLHO to generate an initialization using RL for $x$ timesteps, and then run HO using $y$ timesteps. After $N$ training episodes, we take the initialization generated by the RL step of RLHO and use it to initialize a HO that will run until convergence.

Table \ref{Optimality binpacking3} and Table \ref{Optimality binpacking4} count the average number of used bins of the best solution during training with $x=128, y=5000, t_m=5$ and $x=128, y=50000, t_m=5$ respectively, over 5 independent trials. We also report results where random perturbations (\emph{Random}) are used instead of RL to generate the initial solutions as a baseline. We collect results for 10000 iterations of running RLHO and Random until convergence.

Our results show that RLHO does learn better initializations for HO than \emph{Random}, and the performance gap increases with larger problem sizes. The training signal provided by the HO performance used to augment the value function indeed does help RLHO allow heuristic algorithms to perform better. 
Most interestingly, when the RL part of RLHO is trained using signal from SA that is run for 5000 steps, the initialization it generates is still effective for SA that runs until convergence, e.g. millions of timesteps.

\begin{table}[H]
\centering
	\begin{tabular}{|c|c|c|} 
		\hline 
		$n$  & RLHO & Random, then HO \\  
		\hline
		100 & \textbf{59}  & 69 \\
		\hline
		200 & \textbf{128.4} & 141 \\
		\hline
		500 & \textbf{347}  & 361\\
		\hline
		1000 & \textbf{714} & 734 \\
		\hline 
	\end{tabular}%
	\caption{Average cost of the best solution found by each algorithm with $y = 5000/50000$ HO steps.}
	\label{Optimality binpacking3}
\end{table}
\vspace{-0.7 in}
\begin{table}[H]
\centering
	\begin{tabular}{|c|c|c|} 
		\hline 
		$n$  & RLHO  & Random, then HO\\ 
		\hline
		100 & \textbf{59}  & 69 \\
		\hline
		200 & \textbf{127} & 141 \\
		\hline
		500 & \textbf{344.4}  & 359\\
		\hline
		1000 & \textbf{711}  & 731 \\
		\hline 
	\end{tabular}%
	\caption{Average cost of the best solution found by each algorithm with $y=50000$ HO steps.}
	\label{Optimality binpacking4}
\end{table}

\vspace{-0.2in}
\subsection{Having RL and HO Work Together}

Now we extend our experimental evaluation to answer the following question: can HO help RL train better? Can running HO after an RL training step help RL explore better states? 

We adjust RLHO to perform alternating optimization on a combinatorial optimization problem. RL will generate a solution, which will then be optimized by HO. RL will then be trained with additional signal from HO. The same solution will then be passed back to RL for continuous optimization. This differs from our previous approach because we do not reset the solution on each episode. The greedy nature of HO will perform hill climbing, allowing RL to see more optimal states throughout training.

\begin{table}[H]
\centering
	\begin{tabular}{|c|c|c|} 
		\hline 
		$n$  & RL & RLHO  \\ 
		\hline
		50 & 22 & \textbf{22}    \\
		\hline
		100 & 50 & \textbf{50}   \\
		\hline
		200 & 102 & \textbf{101}   \\
		\hline
		500 & 283 & \textbf{266}   \\
		\hline
		1000 & 613& \textbf{601}  \\
		\hline 
	\end{tabular}%
	\caption{Average cost of the best solution found by each algorithm with $x=128, y=1000$}
	\label{Optimality binpacking1}
\end{table}
\vspace{-0.2in}
\begin{table}[H]
\centering
	\begin{tabular}{|c|c|c|} 
		\hline 
		$n$  & RL & RLHO \\ 
		\hline
		50 & 22 & \textbf{22}   \\
		\hline
		100 & 50 & \textbf{50}    \\
		\hline
		200 & 102 & \textbf{101}   \\
		\hline
		500 & 283 & \textbf{265}   \\
		\hline
		1000 & 613& \textbf{572}  \\
		\hline 
	\end{tabular}%
	\caption{Average cost of the best solution found by each algorithm with $x=128, y=5000$}
	\label{Optimality binpacking2}
\end{table}

\begin{figure}[!h]
    \centering
            \includegraphics[scale=0.5]{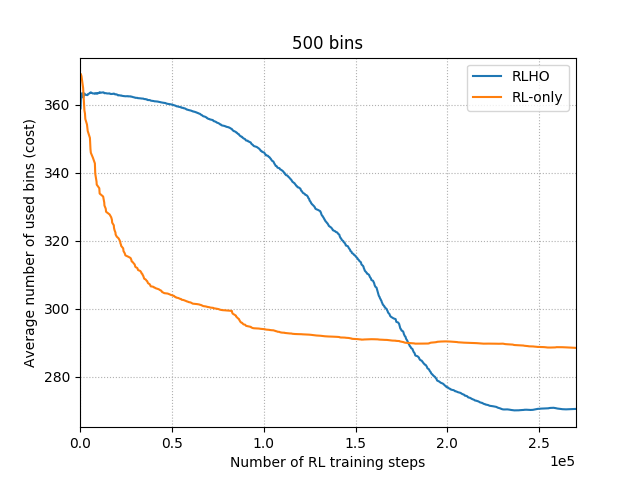}
    \caption{Training performance on 500bins.}
    \label{fig: training}
\end{figure}

We run the two algorithms side-by-side to evaluate our approach. Table \ref{Optimality binpacking1} and Table \ref{Optimality binpacking2} show the average number of used bins of the best solution (over 5 independent runs) searched by both algorithms during training. For \emph{RL}, we simply keep running PPO without any SA. In \emph{RLHO}, PPO learns from SA. We choose to set $x=128$, $y=1000$ and the initial temperature of SA to be $5$. We compare the performance of two algorithms in terms of the number of steps the RL policy performs, with the hyperparameters of the RL part of both approaches kept constant. We also evaluate our approaches on different sizes of the bin packing problem. We report the results until 2000 iterations run for the alternating optimization.

 The convergence curves of all approaches are shown in Figure \ref{fig: training}. We conclude that the pure RL algorithm is more sample efficient but performs worse as the RL algorithm has no additional outlet for exploration. RLHO achieves better performance because it adopts the HO to perform better exploration. 

\section{Conclusion}
In this paper, we propose a novel Reinforcement Learning Driven Heuristic Optimization framework that applies reinforcement learning to learn better initialization for heuristic optimization algorithms. We present the RLHO learning algorithm which builds upon Proximal Policy Optimization and Simulated Annealing. Experimental results on the bin packing problem show that the RLHO learning algorithm does indeed learn better initialization for heuristic optimization, outperforming pure reinforcement learning algorithms. Our approach can be applied towards combinatorial optimization problems that have real world applications.

We hope to further evaluate our methodology on a broad range of other combinatorial optimization problems such as TSP, graph partitioning, and integer programming, with other heuristic algorithms such as evolutionary strategies to demonstrate the power of our approach. We also plan on providing a better and theoretically motivated estimator of heuristic performance to the reinforcement learning agent.

\bibliographystyle{ACM-Reference-Format}
\bibliography{sample-sigconf}


\end{document}